\documentclass[10pt,twocolumn,letterpaper]{article}

\usepackage[pagenumbers]{cvpr} 

\usepackage{graphicx}
\usepackage{amsmath}
\usepackage{amssymb}
\usepackage{booktabs}

\usepackage{multirow}
\usepackage{bbm}
\makeatletter
\newcommand{\figcaption}{\def\@captype{figure}\caption}
\newcommand{\tabcaption}{\def\@captype{table}\caption}
\makeatother
\usepackage[pagebackref,breaklinks,colorlinks]{hyperref}

\usepackage[capitalize]{cleveref}
\crefname{section}{Sec.}{Secs.}
\Crefname{section}{Section}{Sections}
\Crefname{table}{Table}{Tables}
\crefname{table}{Tab.}{Tabs.}
\newcommand{\red}[1]{\textcolor{red}{#1}}
\newcommand{\lightgreen}[1]{\textcolor[RGB]{0,205,102}{#1}}

\def\cvprPaperID{5348} 
\def\confName{CVPR}
\def\confYear{2022}

\begin{document}

\title{CREAM: Weakly Supervised Object Localization via \\ 
Class RE-Activation Mapping}

\author{Jilan Xu$^{1}$ ~Junlin Hou$^{1}$ ~Yuejie Zhang$^{1}$\thanks{Corresponding author} ~Rui Feng$^{1*}$ ~Rui-Wei Zhao$^{2}$\\Tao Zhang$^{3}$ ~Xuequan Lu$^{4}$ ~Shang Gao$^{4}$ 
\\
$^1$School of Computer Science, Shanghai Key Lab of Intelligent Information Processing, \\Shanghai Collaborative Innovation Center of Intelligent Visual Computing, Fudan University\\
$^2$Academy for Engineer and Technology, Fudan University\\
$^3$Shanghai University of Finance and Economics ~$^4$Deakin University \\
{\tt\small \{jilanxu18,jlhou18,yjzhang,fengrui,rwzhao\}@fudan.edu.cn, taozhang@mail.shufe.edu.cn,}\\
{\tt\small \{xuequan.lu,shang\}@deakin.edu.au}
}

\maketitle



\begin{abstract}
Weakly Supervised Object Localization (WSOL) aims to localize objects with image-level supervision. Existing works mainly rely on Class Activation Mapping (CAM) derived from a classification model. However, CAM-based methods usually focus on the most discriminative parts of an object (i.e., incomplete localization problem). In this paper, we empirically prove that this problem is associated with the mixup of the activation values between less discriminative foreground regions and the background. To address it, we propose Class RE-Activation Mapping (CREAM), a novel clustering-based approach to boost the activation values of the integral object regions. To this end, we introduce class-specific foreground and background context embeddings as cluster centroids. A CAM-guided momentum preservation strategy is developed to learn the context embeddings during training. At the inference stage, the re-activation mapping is formulated as a parameter estimation problem under Gaussian Mixture Model, which can be solved by deriving an unsupervised Expectation-Maximization based soft-clustering algorithm. By simply integrating CREAM into various WSOL approaches, our method significantly improves their performance. CREAM achieves the state-of-the-art performance on \textit{CUB}, \textit{ILSVRC} and \textit{OpenImages} benchmark datasets. Code will be available at https://github.com/Jazzcharles/CREAM.
\end{abstract}

\begin{figure}[t]
\centerline{\includegraphics[width=\columnwidth]{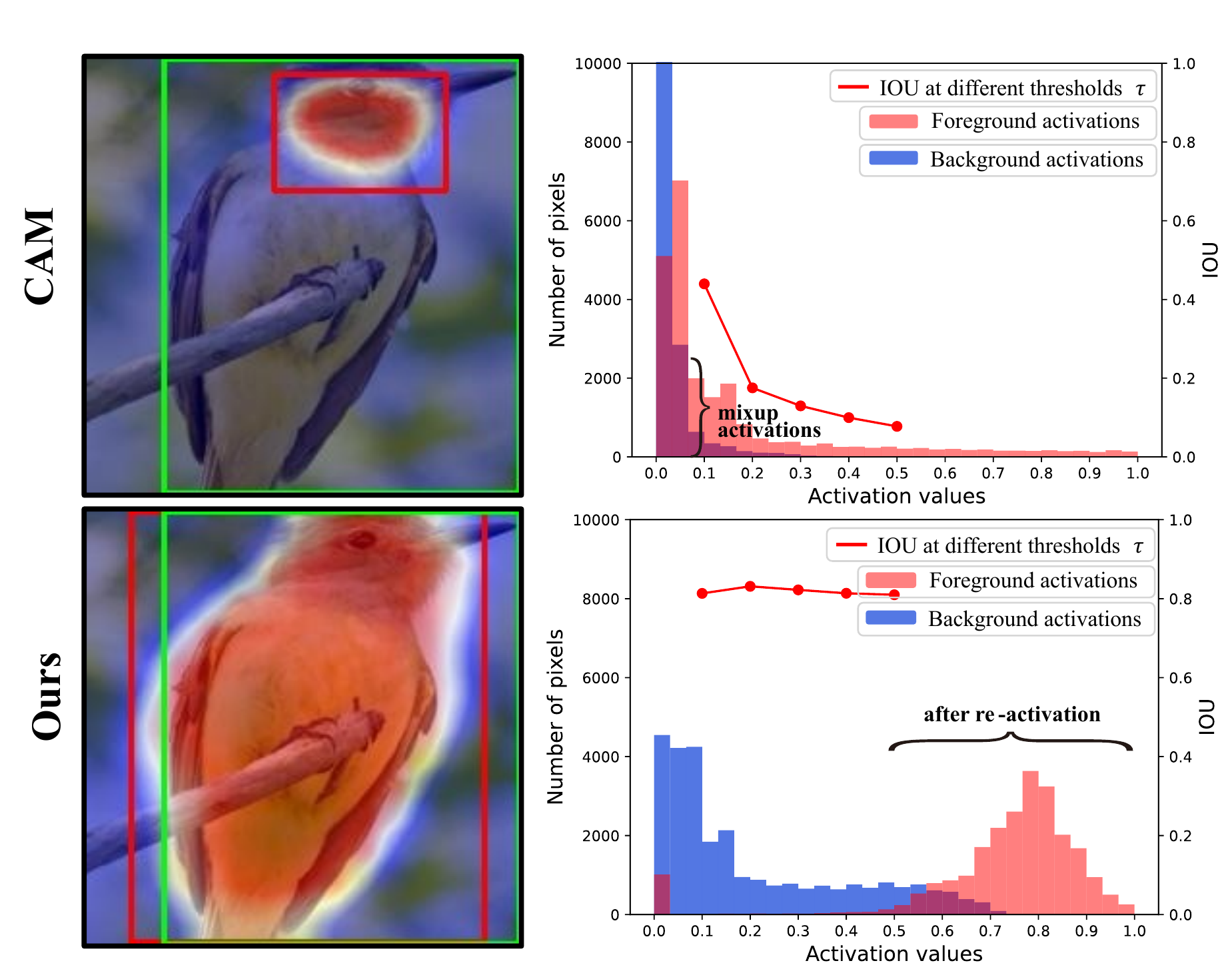}}
\caption{Histograms of the activation values in CAM and our proposed CREAM. The red curves show the IOU between the ground-truth box and the predicted box when $\tau$ varies from 0.1 to 0.5.}
\label{hist}
\vspace{-0.3cm}
\end{figure}

\section{Introduction}
\label{sec:intro}
Weakly Supervised Object Localization (WSOL) aims to localize objects only belonging to one class in each image using image-level supervision \cite{cam,acol,attentiondropout,zhang2020PSOL}. 
WSOL alleviates massive efforts in obtaining fine annotations. 

Prior works mainly follow the pipeline of training a classification network then deriving the Class Activation Mapping (CAM) \cite{cam} for WSOL. 
The foreground region is determined by the CAM values larger than the threshold. However, CAM only highlights the most discriminative regions (i.e., incomplete localization). Existing works seek to discover the complete object via adversarial erasing \cite{singh2017hide,acol}, spatial regularization \cite{lu2020geometry,SPA} or attention mechanism \cite{attentiondropout}. Most of them still acquire the activation maps via a classification pipeline like CAM to answer ``\textbf{which pixels contribute to the class prediction}''. 

In this paper, we argue that the incomplete localization problem of CAM is associated with the mixup of the activation values between less discriminative foreground regions and the background. Figure \ref{hist} (top) illustrates the distributions of foreground activations (red) and background activations (blue) in CAM. High activations are only dominated by the most discriminative parts (e.g., bird head). In contrast, numerous low activations belong to both the less discriminative parts (e.g., bird body) and the background, making them hard to distinguish. The mixup activations bring the challenge of balancing the precision and the recall of the foreground region. Specifically, the widely adopted threshold $\tau$=0.2 \cite{cam,singh2017hide,attentiondropout} appears too large to completely cover foreground object. Whereas tuning a very small threshold (e.g., $\tau$\textless0.1) may induce significant false-positive localizations. 
Moreover, due to mixup activations, a slight change to the threshold leads to a drastic change to the IOU when $\tau$ is small (red curve), and therefore the localization result is sensitive to the threshold.
Choe et al. \cite{choe2020evaluating} also revealed that a well chosen threshold could lead to a misconception of improvement over CAM. This indicates the infeasibility of improving CAM by simply controlling the threshold.

To address the above challenges, we propose Class RE-Activation Mapping (CREAM), a novel framework for WSOL. We rethink the WSOL task by answering ``\textbf{whether the pixel is more similar to the foreground or the background}''. In particular, we introduce class-specific foreground (background) context embeddings describing the common foreground (background) features. During training, a momentum preservation strategy is developed to update the context embeddings under the guidance of CAM. With abundant context information, the learned embeddings serve as initial foreground (background) cluster centroids for each class. During inference, we cast the re-activation mapping as a parameter estimation problem formulated under the Gaussian Mixture Model (GMM) framework. We solve it by deriving an Expectation-Maximization (EM) based soft-clustering algorithm. CREAM boosts the activation values of the integral object, eases the foreground-background separation and shows robustness to the threshold, as shown in Figure \ref{hist} (bottom). It outperforms prior WSOL works on \textit{CUB}, \textit{ILSVRC} and \textit{OpenImages} datasets.

To sum up, the contributions of this paper are as follows: 
\begin{itemize}
\item We propose CREAM, a clustering-based method, to solve the mixup of activations between less discriminative foreground regions and the background by boosting the activations of the integral object regions.
\item We devise a CAM-guided momentum preservation strategy to learn the class-specific context embeddings and use them as initial cluster centroids for re-activation mapping. 
\item We regard re-activation mapping as a parameter estimation problem under the GMM framework and solve it by deriving an EM-based soft-clustering algorithm. 
\item CREAM achieves state-of-the-art localization performance on \textit{CUB}, \textit{ILSVRC} and \textit{OpenImages} benchmark datasets. It can also serve as a plug-and-play method in a variety of existing WSOL approaches.
\end{itemize}

\section{Related Work}
\subsection{Weakly supervised object localization}

\textbf{Regressor-free WSOL methods.} Following Class Activation Mapping (CAM) \cite{cam}, most approaches obtain the class prediction and the localization map with a classification network only \cite{singh2017hide,wei2017object,zhang2018selfproduced,sem,attentiondropout,fcam}. 
To tackle the incomplete localization problem of CAM, several methods adopted an iterative erasing strategy on input images \cite{singh2017hide,yun2019cutmix} or feature maps \cite{wei2017object,zhang2018selfproduced} to force the network's attention to the remaining parts of an object. 
Inspired by the attention mechanism, Zhang et al. \cite{zhang2020inter} leveraged pixel-level similarities across different objects to acquire their consistent feature representations. 
RCAM \cite{bae2020rethinking} offered thresholded average pooling, negative weight clamping and percentile thresholding as ways to improving CAM. SEM \cite{sem} sampled top-K activations as foreground seeds and used them to assign the label for each pixel. Xie et al.\cite{xie2021online} generated compact activation maps under the guidance of low-level features. 
Different from these approaches, our method obtains the activation map via an EM-based soft clustering mechanism, which naturally relieves the incomplete localization problem. 

\textbf{Regressor-based WSOL methods.} This category of work \cite{zhang2020PSOL,lu2020geometry,wei2021shallow,guo2021strengthen} disentangles the WSOL task into image classification and class-agnostic object localization. The intuition is that a model's high localization performance is accompanied by low classification accuracy in early epochs, and vice versa in late epochs. The disentanglement aims to boost both the localization and classification performance. 
PSOL \cite{zhang2020PSOL} first proved that training an additional regression network for bounding box prediction yielded great improvement. The bounding box annotations were obtained using an unsupervised co-localization approach \cite{ddt}. Lu et al. \cite{lu2020geometry} applied a classifier, a generator and a regressor to impose the geometry constraint for compact object discovery.
SLT-Net \cite{guo2021strengthen} produced robust localization results with semantic and visual stimuli tolerance strengthening mechanisms. With the assistance of a separate regression network, these works achieve superior localization performance over most regressor-free methods. Our CREAM is applicable to both the regressor-free and regressor-based methods.

\begin{figure*}[!t]
\centerline{\includegraphics[width=1.0\textwidth]{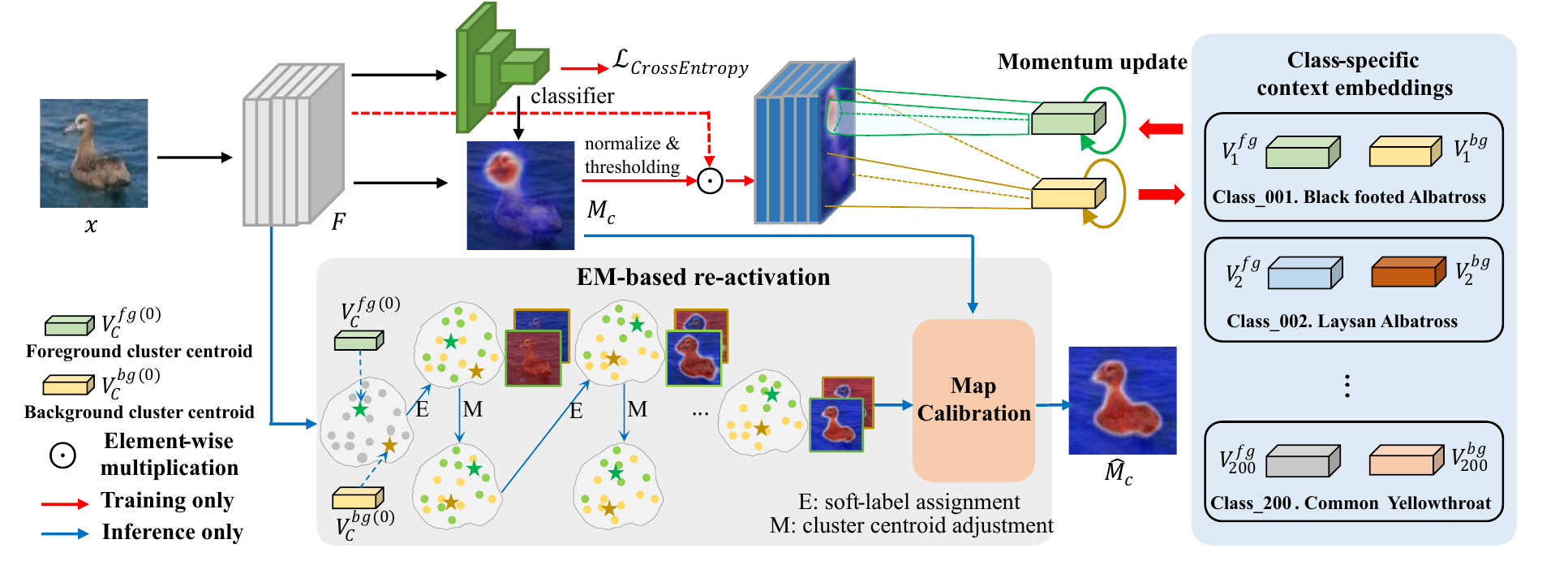}}
\caption{The overall architecture of CREAM. The class-specific context embeddings are maintained during training. At the inference stage, the trained embeddings act as initial cluster centroids and are utilized to re-activate CAM by executing E-step and M-step alternately. Map calibration is conducted to generate the final class re-activation mapping $\hat{M}_c$. Best viewed in color.}
\label{model}
\end{figure*}

\subsection{EM-based deep learning methods}
Expectation-Maximization (EM) algorithm has been frequently adopted in deep learning in recent works \cite{capsule,yang2020prototype,li2019expectation,luo2020weakly,biggs2020left,yan2021framework}. Hinton et al. \cite{capsule} introduced EM routing to group capsules for part-whole relationship construction. Yang et al. \cite{yang2020prototype} designed prototype mixture models for few-shot segmentation. They applied EM algorithm to estimate the models' mean vectors for query images. 
Biggs et al. \cite{biggs2020left} used EM algorithm to learn a 3D shape prior for animal reconstruction. Most prior works jointly optimized the parameters in the EM algorithm and the network's parameters during training. In comparison, we derive an EM based algorithm during inference as an unsupervised soft-clustering mechanism for  foreground-background separation.

\section{Methodology}
\subsection{Revisiting class activation mapping for WSOL}
Let $F \in \mathbb{R}^{d\times h\times w}$ be the feature map in the last convolutional layer, and each $f_k$ corresponds to the feature map at channel $k$. $w_k^c$ is the weight for the $k^{\text{th}}$ channel with regard to class $c$. The class activation mapping $M_c$ is defined as:
\begin{equation}
    M_c = \sum_k w_k^c f_k.   
\end{equation}
Meanwhile, the class prediction can be re-written as:
\begin{equation}
    S_c = \sum_k w_k^c \sum_{i,j} f_k(i,j) = \sum_{i,j} M_c(i,j).
\label{classpred}
\end{equation}
where $i$ and $j$ stand for the spatial location. In this way, solving WSOL with CAM can be interpreted as answering ``which pixels contribute to the class prediction''.

To produce the final bounding box/mask for WSOL evaluation, $M_c$ is then normalized to [0, 1] and thresholded by a pre-defined hyperparameter $\tau$. Only the CAM values greater than $\tau$ are considered as foreground pixels. However, the activation values of the less discriminative foreground region and the background are indistinguishable, as shown in Figure \ref{hist} (top). CAM is thus prone to the incomplete localization problem. 


\subsection{Class re-activation mapping}
For each image, our goal is to discover its foreground region by mapping each pixel into foreground or background cluster. We start by introducing $V^{fg}$ and $V^{bg}$ as the context embeddings. $V^{fg}_c\in\mathbb{R}^{d}$ and $V^{bg}_c\in\mathbb{R}^{d}$ represent the common foreground and background context features in class $c$, respectively. They can also be regarded as two cluster centroids for each class. The ways to learning the embeddings and realizing re-activation based on the learned embeddings are described below. Figure \ref{model} shows the framework of our proposed \textit{Class RE-Activation Mapping}. 

\textbf{Training stage: context embedding learning.} 
We follow exactly the same training procedure and the cross-entropy loss as CAM \cite{cam} except that we maintain the context embeddings additionally. The basic idea of context embedding learning is to use the foreground (background) features of current mini-batch images to update the embeddings. First, we perform hard-thresholding on CAM to obtain the one-hot binary masks $M^{fg}_c$,$M^{bg}_c\in \{0,1\}^{h\times w}$ as foreground-background indicators, which are calculated as:
\begin{equation}
    M^{fg}_c = \mathbbm{1}(M_c \geq \delta), \quad
    M^{bg}_c = \mathbbm{1}(M_c < \delta),
\end{equation}
\noindent where $\mathbbm{1}()$ is an indicator function; and $\delta$ is set to $0.2 \times max(M_c)$ as suggested in \cite{cam}. The foreground (background) features can be retrieved by the element-wise multiplication of the original feature $F$ and the mask $M^{fg}_c (M^{bg}_c)$. 
For each sample $(x,c)$, we exploit a momentum preservation strategy on the foreground (background) embeddings using the spatial average of the foreground (background) features. Suppose $m\in\{fg,bg\}$, the update of the embeddings with regard to class $c$ is given by:
\begin{equation}
    V^{m}_c = \lambda V^{m}_c + (1-\lambda)\frac{1}{||M_c^{m}||_{0}}\sum_{i=1}^{h}\sum_{j=1}^{w} F_{ij}(M^{m}_c)_{ij}, 
\label{update}
\end{equation}
where $F_{ij}$ is the $F$ value at location $(i,j)$; $\lambda$ is the momentum coefficient; $||\cdot||_{0}$ counts the number of non-zero elements. With rich context features, the trained $V^{fg}$ and $V^{bg}$ act as the initial cluster centroids for re-activation.

\textbf{Inference stage: re-activation mapping.} 
At the inference stage, we formulate re-activation mapping as a parameter estimation problem under Gaussian Mixture Model (GMM) \cite{GMM} and solve it using an unsupervised Expectation-Maximization (EM) algorithm \cite{dempster1977maximum}.
EM algorithm is a generalization of Maximum Likelihood Estimation for probabilistic models with latent variables \cite{bishop2006pattern}. 

\noindent\textbf{Problem Formulation.} 
For each sample $x$, the log-likelihood we aim to maximize is given by:  
\begin{equation}
    \log p(x|\theta) = \sum_{i=1}^{h} \sum_{j=1}^{w} \log p(x_{ij} | \theta),
\end{equation}
where $\theta=\{a^{fg},a^{bg},V^{fg}_c,V^{bg}_c\}$ is the model parameter. In particular, we define the model for each pixel $x_{ij}$ as a probability mixture model of two distributions, i.e., foreground distribution and background distribution:
\begin{equation}
    p(x_{ij}|\theta) = \sum_{m\in\{fg,bg\}} a^{m} p^{m}(x_{ij}|V^{m}_c), 
\end{equation}
where the mixing weights $a^{fg},a^{bg}\in [0, 1]$ and $a^{fg}$+$a^{bg}$=1. The foreground (background) base model $p^{fg}(p^{bg})$ is designed to measure the similarity between the image features and the learned embeddings. Instead of using the RBF kernel in GMM, the choice of base models is Gaussian function based on cosine similarity for implementation efficiency:
\begin{equation}
    p^{m}(x_{ij}|V_{c}^{m}) = e^{(V^{m}_{c})^{T} F_{ij} / \sigma}, \quad m\in\{fg,bg\},
\label{similarity}
\end{equation}
where $\sigma$ is a scale parameter. Next, we describe the application of EM in solving the mixture model. We define $Z^{fg},Z^{bg}$$\in[0, 1]^{h\times w}$ as the latent variables. $Z^{fg}_{ij}$ represents the probability of belonging to foreground at location $(i,j)$. 

\noindent\textbf{E-step.} In the E-step, current parameters are utilized to evaluate the posterior distribution of the latent variables, i.e., $p(Z^{fg}|x,a^{fg},V^{fg}_{c})$ and $p(Z^{bg}|x,a^{bg}, V^{bg}_{c})$. In each iteration $t (1\leq t\leq T)$, assuming the model parameters are fixed, the latent variables are computed as:
\begin{equation}
    Z^{m(t)}_{ij} = \frac{a^{m} p^{m}(x_{ij}|V^{m(t)}_c)}{\sum_{m'\in\{fg,bg\}} a^{m'}p^{m'}(x_{ij}|V^{m'(t)}_c)}, m\in\{fg,bg\}.
\label{e-step}
\end{equation}
From the perspective of soft clustering, Eq. \eqref{e-step} calculates the similarity between each pixel feature and context embeddings (i.e., centroids), and assigns a soft label ($fg/bg$) to each pixel. Different from the random initialization in EM, in our case, the initial embeddings have gained sufficient class-specific features from context embedding learning.

\noindent\textbf{M-step.} In the M-step, the purpose is to adjust the context embeddings by maximizing the expected log-likelihood of the image features using the computed latent variables. This enables class-specific context embeddings to be image-specific. The new model parameters can be obtained by:
\begin{equation}
    V^{m(t+1)}_c = \frac{\sum_{i}^{h} \sum_{j}^{w} Z^{m(t)}_{ij} F_{ij}}{\sum_{i}^{h} \sum_{j}^{w} Z^{m(t)}_{ij}}, a^{m(t+1)}=\frac{\sum_{i}^{h} \sum_{j}^{w}Z^{m(t)}_{ij}}{h\times w},
\label{m-step}
\end{equation}
where $m\in\{fg,bg\}$. $V^{fg}_c$ and $a^{fg}$ are updated by the weighted mean of the features and the effective number of pixels assigned to the foreground, respectively. 

Eqs. \eqref{e-step} and \eqref{m-step} execute alternately until convergence. $Z^{fg(T)}$ and $Z^{bg(T)}$ successfully mark the entire object regions. CREAM naturally re-activates the values of the integral object regions by assigning a higher probability of being foreground to them. The consequent benefit of gaining the activation map via feature clustering is that it avoids Global Average Pooling layer's bias towards small areas \cite{bae2020rethinking}.

\subsection{Map calibration}
So far, $Z^{fg(T)}$ and $Z^{bg(T)}$ have been serving as object localization results. According to the underlying assumption of CAM, foreground objects correspond to high values in the activation map. However, we cannot tell from Eqs. \eqref{similarity} and \eqref{e-step} whether higher or lower values correspond to the foreground regions in $Z^{fg(T)}$. To be specific, $Z^{fg(T)}$ is expected to be the final localization map only if its foreground regions have higher values. Otherwise, $Z^{bg(T)}$ should be chosen. To deal with this uncertainty, we perform map calibration by leveraging their averaged probabilities of being the foreground, i.e., $\overline{Z}^{fg}$ and $\overline{Z}^{bg}$ as:
\begin{equation}
    \overline{Z}^{m} = \frac{1}{||M^{fg}_c||_{0}} \sum_{i=1}^{h} \sum_{j=1}^{w}  Z^{m(T)}_{ij} (M^{fg}_c)_{ij},  \quad m\in\{fg,bg\},
\end{equation}
where the foreground region $M^{fg}_c$ can be similarly retrieved under the guidance of CAM as to the training stage. The calibrated foreground map $M_c^{cal}$ is defined as the one with greater averaged probability:
\begin{equation}
    M_c^{cal} = 
    (Z^{fg(T)})^{\mathbbm{1}( \overline{Z}^{fg}\geq \overline{Z}^{bg})}
    (Z^{bg(T)})^{\mathbbm{1}( \overline{Z}^{fg}< \overline{Z}^{bg})}.
\end{equation}


\noindent To further reduce the re-activations over background regions, the final re-activation map $\hat{M}_c$ is formulated as the sum of the calibrated foreground map and CAM:
\begin{equation}
    \hat{M}_c = M_c^{cal} + M_c.
\label{finalcrm}
\end{equation}


\begin{table*}[t]
\centering
\resizebox{\textwidth}{!}{ 
  \begin{tabular}{lccccccccccccc}
    \toprule
    \multirow{2}{*}{Methods}& \multicolumn{4}{c}{CUB (MaxBoxAccV2)} & \multicolumn{4}{c}{ILSVRC (MaxBoxAccV2)} & \multicolumn{4}{c}{OpenImages (PxAP)} & Total \\
    & VGG & Inception & ResNet & Mean & VGG & Inception & ResNet & Mean & VGG & Inception & ResNet & Mean & Mean \\
    \cmidrule(r){1-1} \cmidrule(r){2-5} \cmidrule(r){6-9} \cmidrule(r){10-13} \cmidrule(r){14-14} 
    Center baseline & 54.4 & 54.4 & 54.4 & 54.4 & 48.9 & 48.9 & 48.9 & 48.9 & 45.8 & 45.8 & 45.8 & 45.8 & 52.3 \\
    
    \cmidrule(r){1-1} \cmidrule(r){2-5} \cmidrule(r){6-9} \cmidrule(r){10-13} \cmidrule(r){14-14} 
    CAM\cite{cam}  & 63.7 & 56.7 & 63.0 & 61.1& 60.0 & 63.4 & 63.7 & 62.4 & 58.3 & 63.2 & 58.5 & 60.0 & 61.2 \\
    HaS\cite{singh2017hide} & \lightgreen{+0.0} & \red{-3.3} & \lightgreen{+1.7} & \red{-0.5}  & \lightgreen{+0.6} & \lightgreen{+0.3} & \red{-0.3} & \lightgreen{+0.2} & \red{-0.2} & \red{-5.1} & \red{-2.6} & \red{-2.6} & \red{-1.0} \\
    
    ACoL\cite{acol} & \red{-6.3} & \red{-0.5} & \lightgreen{+3.5} & \red{-1.1} & \red{-2.6} & \lightgreen{+0.3} & \red{-1.4} & \red{-1.2} & \red{-4.0} & \red{-6.0} & \red{-1.2} & \red{-3.7} & \red{-2.0} \\
    
    SPG\cite{zhang2018selfproduced} & \red{-7.4} & \red{-0.8} & \red{-2.6} & \red{-3.6} & \red{-0.1} & \red{-0.1} & \red{-0.4} & \red{-0.2} & \lightgreen{+0.0} & \red{-0.9} & \red{-1.8} & \red{-0.9} & \red{-1.6} \\
    
    ADL\cite{attentiondropout} & \lightgreen{+2.6} & \lightgreen{+2.1} & \red{-4.6} & \lightgreen{+0.0} & \red{-0.2} & \red{-2.0} & \lightgreen{+0.0} & \red{-0.7} &  \lightgreen{+0.4} & \red{-6.4} & \red{-3.3} & \red{-3.1} & \red{-1.3} \\
    
    CutMix\cite{yun2019cutmix} & \red{-1.4} & \lightgreen{+0.8} & \red{-0.2} & \red{-0.3} & \red{-0.6} & \lightgreen{+0.5} & \red{-0.4} & \red{-0.2} & \red{-0.2} & \red{-0.7} & \red{-0.8} & \red{-0.6} & \red{-0.3} \\
    
    RCAM\cite{bae2020rethinking} & - & - & - & - & - & - & - & - & \lightgreen{+1.3} &  \lightgreen{+0.1} & \lightgreen{+2.4} & \lightgreen{+1.3} & -\\
    
    CAM\_IVR\cite{kim2021normalization} &  \lightgreen{+1.5} &  \lightgreen{+4.1} & \lightgreen{+3.9} & \lightgreen{+3.1} & \lightgreen{+1.5} & \lightgreen{+2.1} & \lightgreen{+1.9} & \lightgreen{+1.8} & \lightgreen{+1.0}  & \lightgreen{+0.4} & \lightgreen{+0.5}  & \lightgreen{+0.6} & 
    \lightgreen{+1.8}  \\
    
    \cmidrule(r){1-1} \cmidrule(r){2-5} \cmidrule(r){6-9} \cmidrule(r){10-13} \cmidrule(r){14-14} 
    $\text{CREAM}_{Base}$ &  \lightgreen{+7.8} & \lightgreen{+7.5} & \lightgreen{+10.5} & \lightgreen{+8.6}  & \lightgreen{+6.2} & \lightgreen{+5.5} & \lightgreen{+3.7} & \lightgreen{+5.1} & \lightgreen{+3.7} &
    \lightgreen{+1.4} &
    \lightgreen{+6.2} &
    \lightgreen{+3.8} & \lightgreen{+5.8}
    \\

    \bottomrule
  \end{tabular}
}
\caption{Quantitative results using threshold-independent metrics. Green (red) numbers denote the absolute increase (decrease) over CAM.}
\label{tableopen}
\end{table*}

\section{Experiments}
\subsection{Experimental setup}
\textbf{Datasets.} We evaluate our method on three popular WSOL benchmark datasets, i.e., Caltech-UCSD-Birds-200-2011 (\textit{CUB}) \cite{cub200}, \textit{ILSVRC} \cite{deng2009imagenet} and \textit{OpenImages} \cite{choe2020evaluating}. \textit{CUB} is a fine-grained bird dataset with 200 species. The training set and testing set contains 5,994 and 5,794 images, respectively. \textit{ILSVRC} contains 1.2 million training images and 50,000 validation images of 1,000 classes. 
Both \textit{CUB} and \textit{ILSVRC} datasets provide bounding boxes for WSOL evaluation. 
\textit{OpenImages} is a recently proposed benchmark. It consists of 29,819, 2,500 and 5,000 images of 100 categories for training, validation and testing, respectively. 

\textbf{Evaluation metrics.}
Following \cite{acol,zhang2020PSOL}, we use three threshold-dependent evaluation metrics: Top-1 localization accuracy (Top-1 Loc), Top-5 localization accuracy (Top-5 Loc) and GT-Known localization accuracy (GT-Known). 
We adopt two recently proposed metrics, MaxBoxAccV2 and Pixel-wise Average Precision (PxAP) \cite{choe2020evaluating}. They directly measure the localization performance regardless of the class prediction and the choice of threshold $\tau$.

\textbf{Implementation details.} We implement our method on three backbones pre-trained on ImageNet \cite{deng2009imagenet}, i.e., VGG16 \cite{vgg}, InceptionV3 \cite{inception} and ResNet50 \cite{resnet}. Following \cite{acol,zhang2018selfproduced,choe2020evaluating}, the input images are resized to 256$\times$256 and randomly cropped to 224$\times$224. We perform random horizontal flip during training. We train 50/6/10 epochs on \textit{CUB}, \textit{ILSVRC} and \textit{OpenImages}, respectively. The initial learning rate is 0.001 and divided by 10 for every 15/2/3 epochs on three datasets, respectively. Notably, we set $10\times$ scale of the  learning rate to the classifier due to its random initialization. The min-max normalization strategy is adopted in all the experiments. The SGD optimizer is used with the batch size of 32. On average, the total iterations $T$=2 is sufficient to cover foreground objects.
By default, $\sigma$=8 and $\lambda$=0.8 are set across all datasets. Specifically, we introduce two variants of CREAM in the experiments:

\begin{figure}[t]
\centerline{\includegraphics[width=\columnwidth]{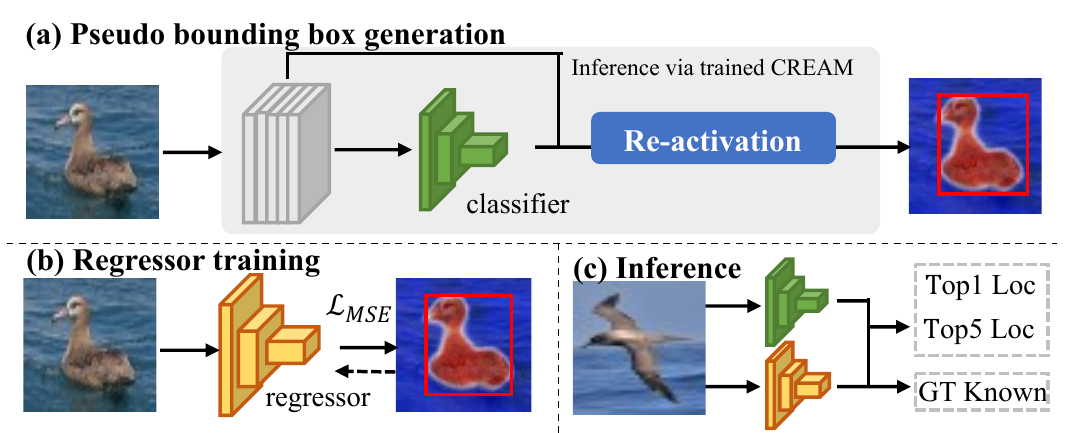}}
\caption{The pipeline of $\text{CREAM}_{Reg}$.}
\label{twostage}
\vspace{-0.3cm}
\end{figure}

\begin{itemize}
    \item $\text{CREAM}_{Base}$ stands for the base CREAM model that only uses the classification network (i.e., regressor-free) to obtain both the class prediction and the localization result, as shown in Figure \ref{model}. Most ablation studies are conducted over $\text{CREAM}_{Base}$.
    \item $\text{CREAM}_{Reg}$ includes a regressor to predict the final bounding box following recent regressor-based methods \cite{zhang2020PSOL,guo2021strengthen}, as illustrated in Figure \ref{twostage}. First, we apply the trained $\text{CREAM}_{Base}$ model to generate the pseudo bounding boxes for the training set. We then train a regressor from scratch using the pseudo bounding boxes as the ground truth. The MSE loss is calculated over four coordinates of the box. During inference, the class prediction and object location are derived from the classifier and the regressor, respectively. By disentangling the WSOL task into classification and class-agnostic object localization, $\text{CREAM}_{Reg}$ solves the contradiction that the localization accuracy gradually decreases while the classification accuracy increases during training \cite{mai2020erasing,zhang2020PSOL}.
\end{itemize}

\begin{table}[t]
\centering
\resizebox{\columnwidth}{!}{
  \begin{tabular}{lcccc}
    \toprule
    \multirow{2}{*}{Methods}& 
    \multirow{2}{*}{Backbone} & 
    \multicolumn{3}{c}{Localization Accuracy} \\
    \cmidrule(r){3-5}
    & & Top-1 Loc. & Top-5 Loc. & GT-Known \\
    \midrule
     CAM \cite{cam} & VGG16& 44.15	& 52.16	& 56.00 \\
     CutMix \cite{yun2019cutmix}  &VGG16& 43.45 & - & - \\
     ACoL \cite{acol} &VGG16& 45.92 & 56.51 & 54.10 \\
     SPG \cite{zhang2018selfproduced} &VGG16& 48.93	& 57.85	& 58.90 \\
     HaS-32 \cite{singh2017hide} &VGG16& 49.50 & - & 71.60 \\
     ADL \cite{attentiondropout} &VGG16& 52.36 & - & 75.40 \\
     DANet \cite{xue2019danet} &VGG16& 52.52 & 61.96 & 67.70 \\
     MEIL \cite{mai2020erasing} &VGG16& 57.46 & - & 73.80\\
     SPA \cite{SPA} &VGG16& 60.27	& 72.50 & 77.29 \\
     RCAM \cite{bae2020rethinking} &VGG16& 61.30 & - & 80.72 \\
     GC-Net \cite{lu2020geometry} &VGG16 & 63.24 & 75.54 & 81.10 \\
     PSOL \cite{zhang2020PSOL} & VGG16& 66.30 & 84.05 & - \\
     SLT-Net \cite{guo2021strengthen} & VGG16 & 67.80 & - & 87.60 \\
     ORNet \cite{xie2021online} & VGG16 & 67.74 & 80.77 & 86.20 \\
     FAM \cite{meng2021foreground} & VGG16 & 69.26 & - & 89.26 \\
     
    \midrule
    $\text{CREAM}_{Reg}$ & VGG16 & \textbf{70.44} & \textbf{85.67} & \textbf{90.98} \\
    \midrule
    \midrule
     DANet \cite{xue2019danet}&InceptionV3& 49.45	& 60.46	& 67.00 \\
     ADL \cite{attentiondropout}&InceptionV3& 53.04 & - & - \\
     SPA \cite{SPA}&InceptionV3& 53.59 & 66.50 & 72.14 \\
     I$^2$C \cite{zhang2020inter}&InceptionV3& 55.99 & 68.34 & 72.60 \\
     PSOL \cite{zhang2020PSOL} &InceptionV3& 65.51 & 83.44 & - \\
     SLT-Net \cite{guo2021strengthen} &InceptionV3& 66.10 & - & 86.50 \\
     FAM \cite{meng2021foreground} & InceptionV3 & 70.67 & - & 87.25 \\
    \midrule
     TS-CAM \cite{gao2021ts}& Deit-S & 71.30 & 83.80 & 87.70 \\    
    \midrule
     $\text{CREAM}_{Reg}$ & InceptionV3 & \textbf{71.76} & \textbf{86.37} & \textbf{90.43} \\
    \bottomrule
  \end{tabular}
}
\caption{Quantitative results on \textit{CUB}.}
\vspace{-0.3cm}
\label{tablecub}
\end{table}

\begin{table}[t]
\centering
\resizebox{\columnwidth}{!}{
  \begin{tabular}{lccccc}
    \toprule
    \multirow{2}{*}{Methods} & 
    \multirow{2}{*}{Backbone} & 
    \multicolumn{3}{c}{Localization Accuracy} \\
    \cmidrule(r){3-5}
    & & Top-1 Loc. & Top-5 Loc. & GT-Known  \\
    \midrule
     CAM\cite{cam} & VGG16& 42.80 & 54.86 & 59.00 \\
     CutMix\cite{yun2019cutmix}   & VGG16& 43.45 & - & - \\
     RCAM\cite{bae2020rethinking}  & VGG16& 44.69 & - & 61.69 \\
     ADL\cite{attentiondropout} & VGG16& 45.92 & - & - \\
     ACoL\cite{acol}  & VGG16& 45.83 & 59.43 & 62.96 \\
     MEIL\cite{mai2020erasing}  & VGG16& 46.81 & - & - \\
     I$^2$C\cite{zhang2020inter} & VGG16& 47.41 & 58.51 & 63.90 \\
     SEM\cite{sem}  & VGG16 & 47.53 & - & 63.74 \\
     SPA\cite{SPA}  & VGG16& 49.56 & 61.32 & 65.05 \\
     PSOL\cite{zhang2020PSOL}  & VGG16& 50.89 & 60.90 & 64.03 \\
     SLT-Net\cite{guo2021strengthen} & VGG16& 51.20 & 62.40 & 67.20 \\
     ORNet\cite{xie2021online}  & VGG16 & 52.05 & 63.94 & 68.27 \\
     FAM\cite{meng2021foreground} & VGG16 & 51.96 & - & \textbf{71.73} \\
    \midrule
     $\text{CREAM}_{Reg}$ & VGG16 & \textbf{52.37} & \textbf{64.20} & 68.32 \\
    \midrule
    \midrule

     CAM\cite{cam}  &InceptionV3& 46.29 &	58.19 & 62.68 \\
     MEIL\cite{mai2020erasing}  &InceptionV3& 49.48 & - & - \\
     GC-Net\cite{lu2020geometry}  & InceptionV3& 49.06 & 58.09 & - \\
     SPA\cite{SPA}  &InceptionV3& 52.73 & 64.27 & 68.33 \\
     SEM\cite{sem}  &InceptionV3& 53.04 & - & \textbf{69.04} \\
     I$^2$C\cite{zhang2020inter}  &InceptionV3& 53.11 & 64.13 & 68.50 \\
     PSOL\cite{zhang2020PSOL} & InceptionV3& 54.82 & 63.25 & 65.21 \\
     SLT-Net\cite{guo2021strengthen} & InceptionV3& 55.70 & 65.40 & 67.60 \\
     FAM\cite{meng2021foreground}  & InceptionV3 & 55.24 & - & 68.62 \\
    \midrule
     TS-CAM\cite{gao2021ts} & Deit-S & 53.40 & 64.30 & 67.60 \\
    \midrule
     $\text{CREAM}_{Reg}$ & InceptionV3 & \textbf{56.07} & \textbf{66.19} & 69.03 \\
    \bottomrule
  \end{tabular}
}
\caption{Quantitative results on \textit{ILSVRC}.}
\label{tableimgnet}
\end{table}

\subsection{Comparison with state-of-the-art methods}



\textbf{Results on CUB.}
Tables \ref{tableopen}, \ref{tablecub} and \ref{resnet} show the quantitative comparisons with other methods. Regarding MaxboxAccV2 in Table \ref{tableopen}, $\text{CREAM}_{Base}$ achieves a consistent improvement on all the backbones, e.g., up to 10.5\% on ResNet compared with vanilla CAM. For GT-Known that measures the localization performance only, our method reaches new state-of-the-art GT-Known performance of 90.98\% and 90.43\% on VGG16 and InceptionV3, respectively. Meanwhile, our method retains the highest accuracy on Top-1 Loc and Top-5 Loc. Especially when the backbone is InceptionV3, $\text{CREAM}_{Reg}$ surpasses PSOL \cite{zhang2020PSOL} by 2.9\% on Top-5 Loc. Despite variations in bird size and view angle, the results show that our method succeeds in localizing objects in the fine-grained dataset \textit{CUB}.

\textbf{Results on ILSVRC.}
Tables \ref{tableopen}, \ref{tableimgnet} and \ref{resnet} contain the results on \textit{ILSVRC}. \textit{ILSVRC} is more challenging than \textit{CUB} because of its 1,000 classes and multiple salient objects in an image, resulting in potential false-positive localizations. Despite the difficulties, our method respectively exceeds CAM by 9.3\% and 6.4\% on VGG16 and InceptionV3. Besides, $\text{CREAM}_{Base}$ outperforms CAM by an average of 5.1\% on MaxBoxAccV2. In terms of Top-1 Loc., our method achieves a modest improvement of 0.3\%, 0.3\% and 1.2\% over the second best methods using VGG16 and InceptionV3 and ResNet50, respectively. 

\textbf{Results on OpenImages.} Regarding pixel-wise WSOL evaluation, $\text{CREAM}_{Base}$ also achieves high localization performance. As listed in Table \ref{tableopen}, when the backbone is ResNet50, $\text{CREAM}_{Base}$ boosts by 6.2\% and 3.8\% over vanilla CAM and the previous best-performed model RCAM \cite{bae2020rethinking}, respectively. We attribute the gain to the pixel-level EM clustering in CREAM. 

The last column in Table \ref{tableopen} shows the total mean scores across different datasets and threshold-independent metrics. $\text{CREAM}_{Base}$ outperforms CAM and CAM\_IVR \cite{kim2021normalization} by 5.8\% and 4.0\% on average. It reveals that the performance gain of $\text{CREAM}_{Base}$ comes from the improved activation map instead of the choice of a better threshold.

     

\begin{table}
\centering
\resizebox{\columnwidth}{!}{ 
  \begin{tabular}{lccccc}
     \toprule
     \multirow{2}{*}{Methods} & \multirow{2}{*}{Backbone}
     &\multicolumn{2}{c}{CUB} &\multicolumn{2}{c}{ILSVRC} \\
     \cmidrule(r){3-4} \cmidrule(r){5-6} 
     & & Top-1 Loc. & GT & Top-1 Loc. & GT \\
     \midrule
      RCAM \cite{bae2020rethinking} & ResNet50 & 59.53 & 77.58 & 49.42 & 62.20 \\
      PSOL \cite{zhang2020PSOL} & ResNet50 & 70.68 & - & 53.98 & 65.44 \\
      FAM \cite{meng2021foreground} & ResNet50 & 73.74 & 85.73 & 54.46 & 64.56 \\
      \midrule
      $\text{CREAM}_{Reg}$ & ResNet50 & \textbf{76.03} & \textbf{89.88} & \textbf{55.66} & \textbf{69.31} \\
     \bottomrule
  \end{tabular}
}
\caption{Quantitative results using ResNet50.}
\vspace{-0.3cm}
\label{resnet}
\end{table}

\subsection{Ablation study}
In Eq. \eqref{finalcrm}, we combine the calibrated foreground map $M_c^{cal}$ with CAM ($M_c$) to get the final re-activation map $\text{CREAM}_{Base}$ at the inference stage. In Table \ref{ablations}, we compare CAM (row 1), calibrated foreground map (row 2), $\text{CREAM}_{Base}$ (row 3), and $\text{CREAM}_{Reg}$ (row 4). The calibrated foreground map outperforms the vanilla CAM by a large margin, 29.8\% and 5.3\% on \textit{CUB} and \textit{ILSVRC} datasets, respectively. It demonstrates the effectiveness of the proposed re-activation mapping. By combining CAM and the calibrated foreground map, $\text{CREAM}_{Base}$ further boosts the localization performance. An improvement of 3.7\% is observed on \textit{ILSVRC}. This is because CAM aids $\text{CREAM}_{Base}$ in suppressing the activation values of the background regions, especially on \textit{ILSVRC} where $\text{CREAM}_{Base}$ has potential false-positive localizations. With the aid of the additional separate regression network, $\text{CREAM}_{Reg}$ consistently improves $\text{CREAM}_{Base}$ on both datasets.

\begin{table}
\centering
\resizebox{\columnwidth}{!}{ 
  \begin{tabular}{lcccc}
     \toprule
     \multirow{2}{*}{Methods}
     &\multicolumn{2}{c}{CUB} &\multicolumn{2}{c}{ILSVRC} \\
     \cmidrule(r){2-3} \cmidrule(r){4-5}
     & Top-1 & GT & Top-1 & GT \\
     
     \midrule
      Baseline CAM ($M_c$) &44.2 &56.0 & 42.8 & 59.0 \\
      Caliberated foreground map ($M_c^{cal}$)& 65.5 &85.8& 49.1 & 64.3 \\
      $\text{CREAM}_{Base} (M_c + M_c^{cal})$ & 67.4 &86.7 &51.8 & 68.0 \\
      $\text{CREAM}_{Reg} (M_c + M_c^{cal} + \text{regressor})$ & 70.4 &91.0 & 52.4 &68.3 \\
     \bottomrule
  \end{tabular}
}
\caption{Ablation study on CREAM using VGG16.}
\label{ablations}
\end{table}

\begin{figure}[t]
\centerline{\includegraphics[width=\columnwidth]{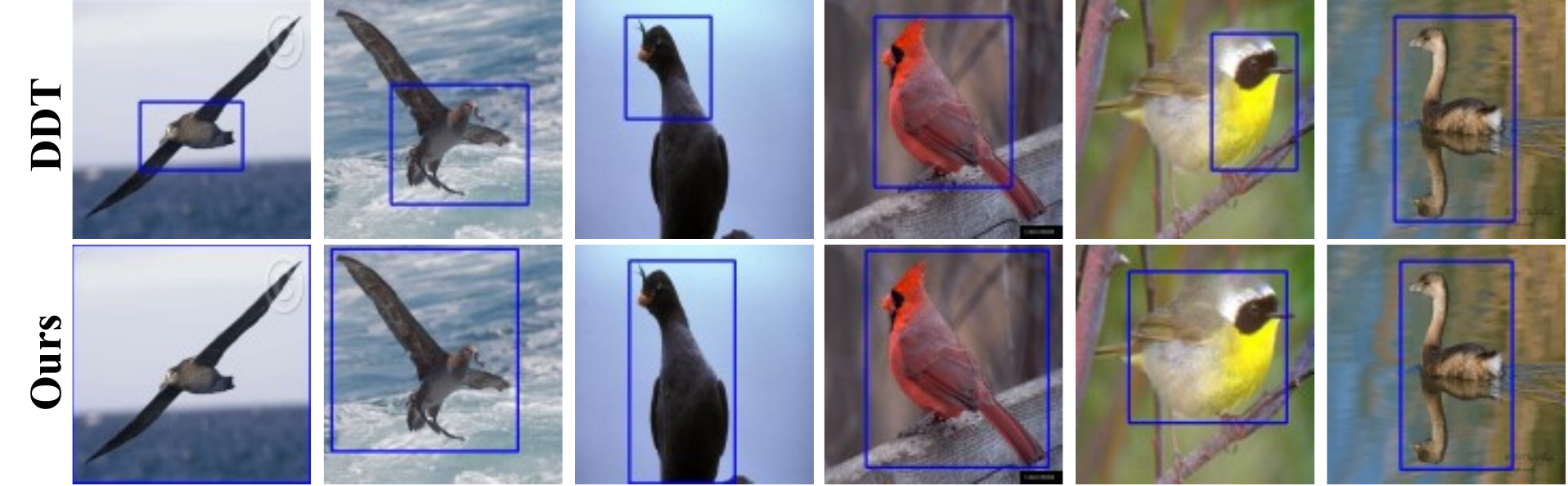}}   
\caption{Comparison of generated pseudo bounding boxes. }
\label{genbox}
\end{figure}

\begin{table}[t]
\centering
\resizebox{\columnwidth}{!}{ 
  \begin{tabular}{lcccccc}
    \toprule
     \multirow{2}{*}{Methods}
     & \multicolumn{3}{c}{CUB} & \multicolumn{3}{c}{ILSVRC} \\
    \cmidrule(r){2-4} \cmidrule(r){5-7} 
    
    & VGG & Inception & ResNet & VGG & Inception & ResNet  \\
    \cmidrule(r){1-1} \cmidrule(r){2-4} \cmidrule(r){5-7}
    
    DDT\cite{ddt} & 84.6 & 51.8 & 72.4 & 61.4 & 51.9 & 59.9  \\
    SLT-Net\cite{guo2021strengthen} & 85.6 & \textbf{78.6} & 68.2 & 63.4 & 65.7 & 54.0  \\
    $\text{CREAM}_{Base}$ & \textbf{86.7} & 72.8 & \textbf{88.0} & \textbf{68.0} & \textbf{70.8} & \textbf{69.1}  \\
    \bottomrule
  \end{tabular}
}
\caption{GT-Known localization performance of pseudo bounding box generation methods using different backbones. }
\label{tablegenbox}
\end{table}

    
    

\section{Discussion}
\textbf{Pseudo bounding box generation.}
In PSOL \cite{zhang2020PSOL}, a co-supervised method DDT \cite{ddt} was used to generate pseudo bounding boxes for the training set before a separate classification-regression model was trained. For comparison, we replace DDT with $\text{CREAM}_{Base}$ for bounding box generation. Following PSOL, we also adopt the resolution of $448\times 448$ when generating boxes. The input image resolution of the regression network remains $224\times 224$. 
As listed in Table \ref{tablegenbox}, $\text{CREAM}_{Base}$ surpasses DDT by a large margin. On ResNet50 backbone, a significant improvement of 15.6\% on \textit{CUB} is observed.
Figure \ref{genbox} shows the examples of the generated pseudo bounding boxes. DDT fails to capture the entire birds when they have long wings or tails, whereas $\text{CREAM}_{Base}$ generates accurate boxes regardless of the changes in bird scale. However, a potential drawback is that both $\text{CREAM}_{Base}$ and DDT have difficulty in dealing with the reflection, as shown in the last column.

The regression networks in prior works directly predict the coordinates of the bounding box without improving the localization map. They have difficulties in fitting pixel-wise WSOL evaluation (e.g., on \textit{OpenImages}). Notably, our proposed $\text{CREAM}_{Base}$ improves the activation map before the pseudo box generation in $\text{CREAM}_{Reg}$, making CREAM applicable to both bounding box and pixel-wise evaluation.

\begin{figure*}[t]
\centerline{\includegraphics[width=\textwidth]{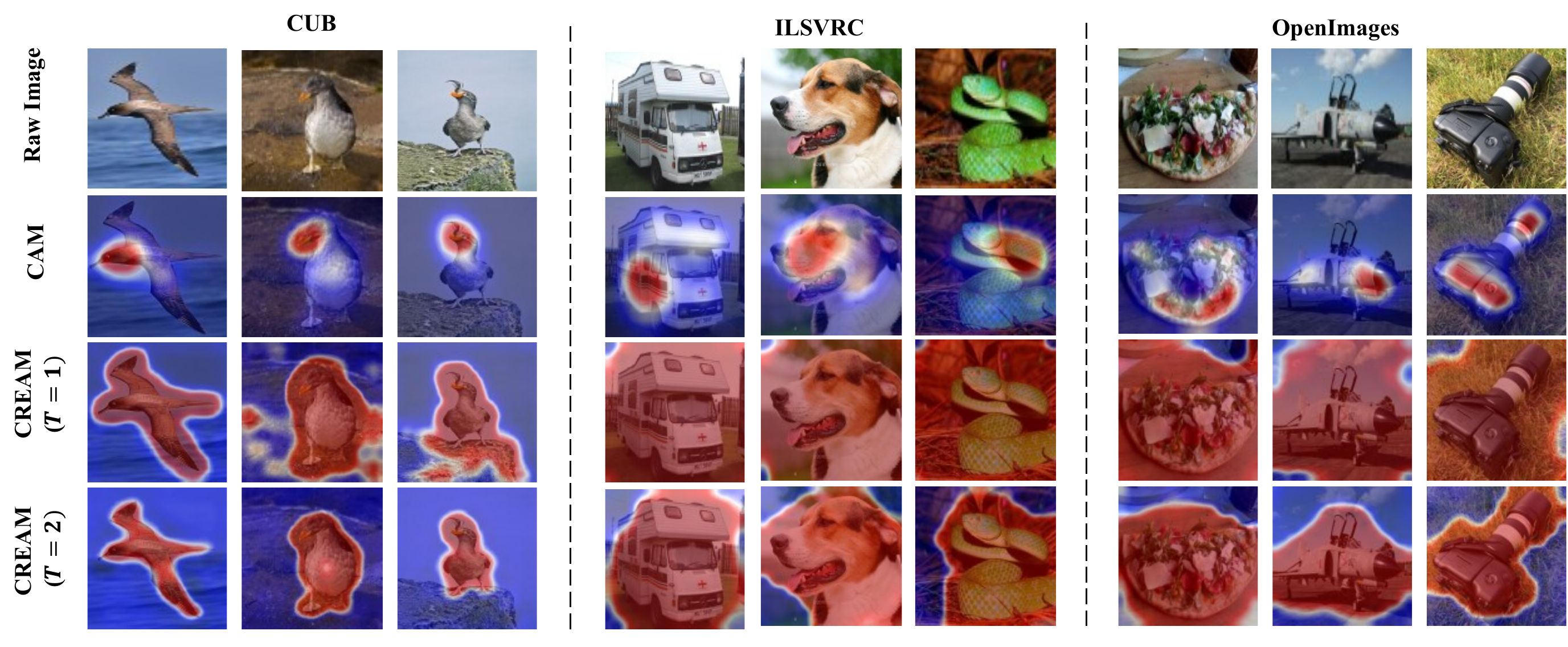}}
\caption{Visualization results of the raw image, CAM, $\text{CREAM}_{Base}$ ($T$=1) and $\text{CREAM}_{Base}$ ($T$=2) on \textit{CUB}, \textit{ILSVRC}, and \textit{OpenImages}. 
}
\label{crm}
\end{figure*}

\textbf{Iterations $T$ and momentum coefficient $\lambda$.}
Tables \ref{iteration} and \ref{momentum} demonstrate the changes of GT-Known with different iterations $T$ and momentum coefficients $\lambda$. Both are conducted on $\text{CREAM}_{Base}$. On both \textit{CUB} and \textit{ILSVRC}, $\text{CREAM}$ yields the best performance when $T$=2. 
CREAM ($T$=2) outperforms CAM by nearly 30\% on \textit{CUB} with only 4.3-ms longer inference time per image. They have similar training time as context embedding learning is not involved in the back propagation.
We visualize the re-activation maps in Figure \ref{crm}. When $T$=1, the re-activation map roughly captures the foreground object, but there are still false-positive pixels. When $T$=2, as E-step and M-step execute in turn, the improved foreground (background) cluster centroids have a more precise indication on foreground (background), and the re-activation map is automatically refined. Nevertheless, we find that the re-activation map tends to oversmooth when $T$ is large. 
As shown in Table \ref{momentum}, $\text{CREAM}$ is not sensitive to the momentum coefficient. A large $\lambda$ only performs slightly better. When $\lambda$=1.0, the context embeddings are equivalent to randomly initialized vectors. In this case, the embeddings include much random noise in the EM process, and the performance of $\text{CREAM}$ drops dramatically. This suggests that preserving rich context information in context embeddings helps improve the localization.

\begin{table}[t]
\begin{minipage}[t]{0.47\columnwidth}
  \centering
  \makeatletter\def\@captype{table}\makeatother
  \resizebox{1.0\columnwidth}{!}{
  \begin{tabular}{cccc}
    \toprule
    $T$ & CUB & ILSVRC & Time (s) \\
    \midrule
    0 & 56.0 & 59.0 & 0.0017 \\
    1 & 69.3 & 65.5 & 0.0058\\
    2 & 86.7 & 68.0 & 0.0060 \\
    3 & 82.2 & 67.0 & 0.0062\\
    4 & 75.3 & 66.4 & 0.0065\\
    \bottomrule
  \end{tabular}}
  \caption{GT-Known and inference time over iterations $T$.}
  \label{iteration}
\end{minipage}
\hfill
\begin{minipage}[t]{0.5\columnwidth}
\centering
\makeatletter\def\@captype{table}\makeatother
\resizebox{0.7\columnwidth}{!}{
  \begin{tabular}{ccc}
    \toprule
    $\lambda$ & CUB & ILSVRC \\
    \midrule
    0.2 & 85.2 & 67.3 \\
    0.6 & 85.4 & 67.7 \\
    0.8 & 86.7 & 68.0 \\
    0.99 & 85.6 & 67.9 \\
    1.0 & 53.1 & 60.3 \\
    \bottomrule
  \end{tabular}}
  \caption{GT-Known over momentum coefficients $\lambda$.}
  \label{momentum}
  \end{minipage}
\end{table}

\textbf{Visualization of the localization results.} As shown in Figure \ref{crm}, our method successfully re-activates complete foreground regions in comparison to CAM. On \textit{CUB}, $\text{CREAM}_{Base}$ even embodies the capability of precise localization by highlighting the objects with a clear contour. We attribute this contour-aware localization to the EM-based re-activation. In EM process, the soft label for each pixel feature is solely determined by the similarity between the feature and the context embeddings. The pixel belongs to the foreground as long as it is more similar to the foreground context embedding than that of the background. 



\begin{figure}[t]
\centerline{\includegraphics[width=\columnwidth]{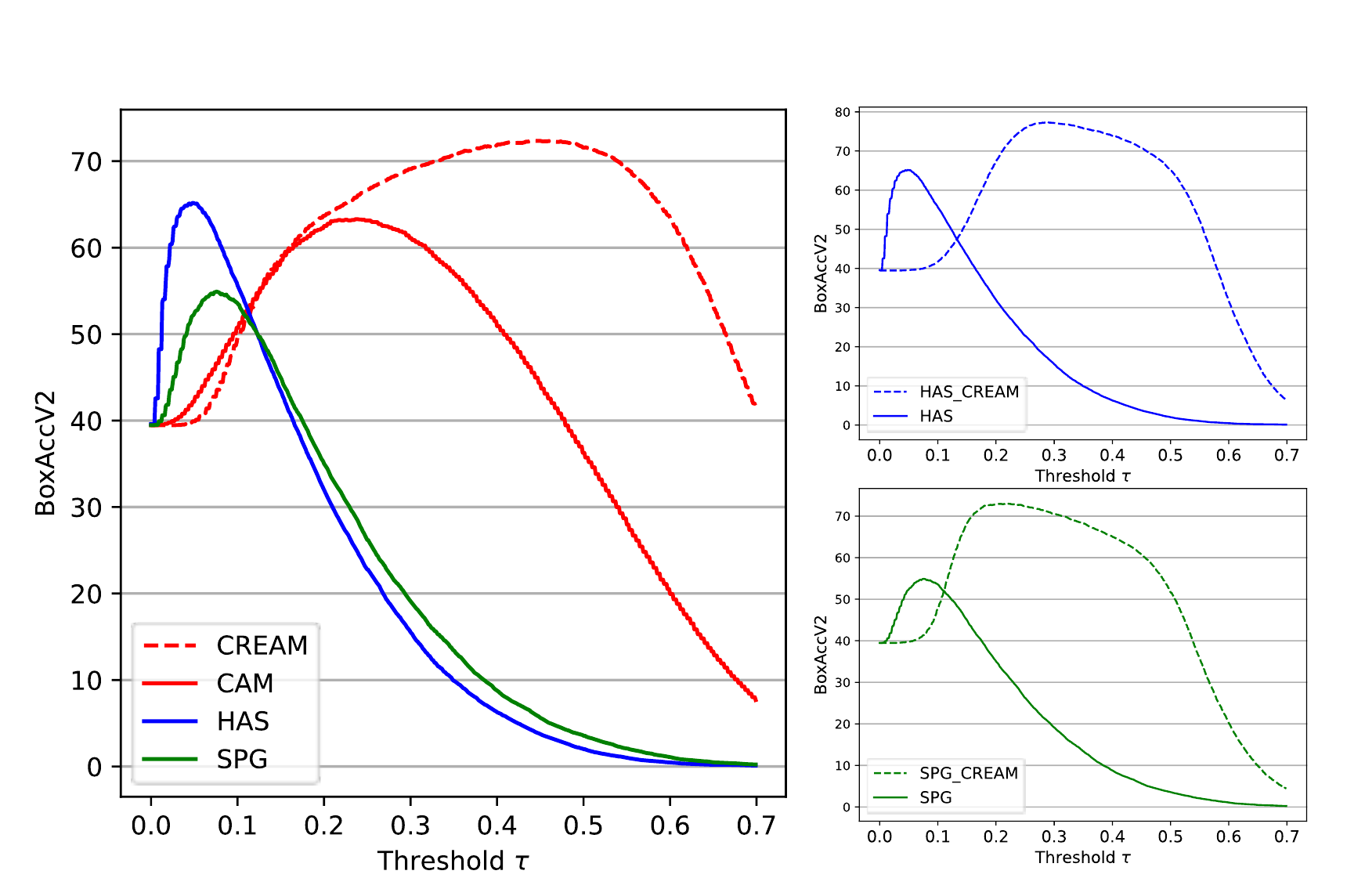}}
\caption{BoxAccV2 under different thresholds on \textit{CUB}. Solid (dashed) lines show the results of baselines (our method). }
\label{curve}
\vspace{-0.3cm}
\end{figure}

\textbf{The effect of re-activation and generalization ability of CREAM.} Following \cite{choe2020evaluating}, we re-implement CAM, HAS \cite{singh2017hide} and SPG \cite{zhang2018selfproduced} on BoxAccV2 \cite{choe2020evaluating} using VGG16. As illustrated in Figure \ref{curve} (left), HAS and SPG perform better than CAM only when $\tau$ is small (e.g. $\tau$\textless0.1). When $\tau$ is large (e.g. $\tau$\textgreater0.2), however, there is no apparent improvement by CAM. In comparison, $\text{CREAM}_{Base}$ curve is above CAM curve for most thresholds. Besides, it is flatter around the peak, showing that $\text{CREAM}_{Base}$ is more robust to the choice of threshold. 


We integrate $\text{CREAM}_{Base}$ into HAS and SPG, resulting in HAS\_$\text{CREAM}_{Base}$ and SPG\_$\text{CREAM}_{Base}$, respectively. As shown in Figure \ref{curve} (right), both HAS\_$\text{CREAM}_{Base}$ and SPG\_$\text{CREAM}_{Base}$ significantly promote the localization performance with an improvement of 10\% to 60\% under different thresholds. The baselines only perform better when $\tau$\textless0.1 as their foreground activations gather around 0.1. Whereas the foreground activations in our CREAM-based counterparts appear larger. Their peaks are also more flatter than original methods. The results demonstrate the effectiveness and the generalization ability of CREAM.


\section{Conclusion}
In this paper, we attribute the incomplete localization problem of CAM to the mixup of the activations between less discriminative foreground regions and the background. We propose CREAM to boost the activation values of the full extent of object. A CAM-guided momentum preservation strategy is exploited to learn the class-specific context embeddings. During inference, re-activation is formulated as a parameter estimation problem and solved by a derived EM-based soft-clustering algorithm. Extensive experiments show that CREAM outperforms CAM substantially and achieves state-of-the-art results on \textit{CUB}, \textit{ILSVRC} and \textit{OpenImages}. CREAM can be easily integrated into various WSOL approaches and boost their performance.

\section*{Acknowledgement}
This work is supported by National Science and Technology Innovation 2030 - Major Project (No.
 2021ZD0114001; No. 2021ZD0114000), National Natural Science Foundation of China (No. 61976057; No. 62172101), the Science and Technology Commission of Shanghai Municipality (No. 21511101000; No. 20511101203), the Science and Technology Major Project of Commission of Science and Technology of Shanghai (No. 2021SHZDZX0103).

\appendix

{\small
\bibliographystyle{ieee_fullname}
\bibliography{main}
}

\end{document}